\documentclass[review]{elsarticle}

\usepackage{booktabs}

\usepackage[ruled]{algorithm2e}
\usepackage{times}
\usepackage{bm}
\usepackage{graphicx}
\usepackage{amsmath}
\usepackage{caption}
\usepackage{booktabs}
\usepackage{multirow}
\usepackage{xcolor}
\usepackage{float}
\usepackage{stfloats}
\usepackage{array}
\usepackage{url}
\usepackage{CJKutf8}
\usepackage{graphicx}  

\usepackage{epstopdf}
\journal{Neurocomputing}

\begin{document}

\begin{frontmatter}

\title{DEIM:An effective deep encoding and interaction model for sentence matching \tnoteref{t}}

\author[mymainaddress]{Kexin Jiang}

\author[mymainaddress]{Yahui Zhao\corref{mycorrespondingauthor}}

\author[mymainaddress]{Rongyi Cui}

\author[mymainaddress]{Zhenguo Zhang}
\ead{\{2020010075,yhzhao,cuirongyi,zgzhang\}@ybu.edu.cn}

\cortext[mycorrespondingauthor]{Corresponding author}

\address[mymainaddress]{Department of Computer Science and Technology, Yanbian University.\\
977 Gongyuan Road, Yanji, P.R.China 133002.}

\begin{abstract}
Natural language sentence matching is the task of comparing two sentences and identifying the relationship between them.It has a wide range of applications in natural language processing tasks such as reading comprehension, question and answer systems. The main approach is to compute the interaction between text representations and sentence pairs through an attention mechanism, which can extract the semantic information between sentence pairs well. However,this kind of method can not gain satisfactory  results when dealing with complex semantic features. To solve this problem, 
we propose a sentence matching method based on deep encoding and interaction to extract deep semantic information. In the encoder layer,we refer to the information of another sentence in the process of encoding a single sentence, and later use a heuristic  algorithm to fuse the information. In the interaction layer, we use a bidirectional attention mechanism and a self-attention mechanism to obtain deep semantic information.Finally, we perform a pooling operation and input it to the MLP for classification. we evaluate our model on three tasks: recognizing textual entailment, paraphrase recognition, and answer selection. We conducted experiments on the SNLI and SciTail datasets for the recognizing textual entailment task, the Quora dataset for the paraphrase recognition task, and the WikiQA dataset for the answer selection task. The experimental results show that the proposed algorithm can effectively extract deep semantic features that verify the effectiveness of the algorithm on sentence matching tasks.
\end{abstract}

\end{frontmatter}

\section{Introduction}
Natural language sentence matching is the task of comparing two sentences and identifying the relationship between them. It is a fundamental technique for various tasks and has been successfully applied in many areas of natural language processing. For example, reading comprehension \cite{sugawara2020assessing}, question and answer systems \cite{liu2020asking}, and machine translation \cite{li2021incorporating}. In machine reading comprehension tasks, the correct answer can be selected by semantic matching between the context and the question. In question and answer system tasks, two directions rely on sentence matching: (1) question retrieval, which determines the match between the query and the existing question;  (2) answer selection, which determines the match between the query and the answer.In machine translation tasks, the effectiveness of the translation can be evaluated by semantic matching between the two languages. Natural language sentence matching tasks include 
recognizing textual entailment\cite{li2020sa}, interpretation recognition\cite{zhang2019deep}, answer selection\cite{wu2020building}, etc. Recognizing Textual Entailment (RTE), proposed by Dagan {\it et al}. \cite{dagan2004probabilistic}, is a study of the relationship between premises and assumptions. it mainly includes entailment, contradiction, and neutrality.The task of paraphrase identification is to determine whether two texts hold the same meaning. Answer selection task is to choose one or more suitable answers for a given question from several candidate answers.

The existing sentence matching models can be grouped into two main categories:  sentence matching models based on traditional methods and sentence matching models based on deep learning. The traditional sentence matching methods mainly rely on manually defined features to calculate the similarity between sentences. Algorithms ,such as TF-IDF and BM25 \cite{robertson2009probabilistic} ,obtain sparse matrices with large dimensions, which are difficult to extract deep semantic information. In recent years, due to the rapid development of deep learning and the release of related large-scale datasets with annotations, such as SNLI \cite{bowman2015large} and MultiNLI \cite{williams2018broad}, deep learning-based methods are receiving increasing attention for sentence matching problems.The main idea is to encode two sentences into vectors by deep learning methods\cite{shen2020learning}\cite{zhou2016learning} or to interact with two sentences using attention mechanisms\cite{yang2019simple}\cite{kim2019semantic}.Deep learning-based methods require a large amount of
 training data to obtain optimized parameters. To save training time,
 pre-trained models have been proposed to train parameters in advance by a large open corpus. ELMo\cite{peters2018deep}  captured the contextual information to adjust the word semantic based on BiLSTM. BERT\cite{devlin2019bert} and RoBERTa\cite{liu2019roberta} use Transformer\cite{vaswani2017attention} as the basic encoder and they achieve good results on multiple tasks in NLP. All these methods can extract sentence semantic information effectively, so their performance is higher than sentence matching based on traditional methods. 

To extract deep semantic information, we propose a sentence matching method based on deep encoding and interaction.Our model is divided into an embedding layer, an encoding layer, an interaction layer and an output layer. In the embedding layer, we use dynamic word vector Elmo and static word vector Glove for word embedding to solve the problem of multiple meanings of a word. In the encoding layer, we use CNN and self-attention mechanism to obtain the sentence representation, and add the residual connection to enrich the semantic information. Then we  use the attention mechanism to interact the two sentences to get the representation of the sentence after the interaction. To  effectively fuse the sentence information before and after the interaction, we propose a heuristic fusion algorithm instead of simple splicing. In the interaction layer module, we use the two-way attention interaction in the machine reading comprehension model to obtain the intrinsic semantic information between two sentences and then use the self-attention mechanism to obtain the deep semantic information. Finally, 
we use multi-layer perceptron output results. We conducted experiments on SNLI and SciTail datasets, Quora dataset, and WikiQA dataset and verify the effectiveness of the method on the sentence matching task.   

\section{Related Work}

The  deep learning-based sentence matching models are divided into two categories; (1)Representation-based sentence matching models, which the  main purpose is to represent a sentence as a vector. (2) Interaction-based sentence matching models,which aim to obtain complex interaction information between sentences.

\subsection{Representation-based sentence matching model}

Representation-based sentence matching models focus on constructing a representation vector of sentences. The traditional methods for sentence matching include the following: similarity-based methods \cite{renhan2015recognizing}, rule-based methods \cite{hu2020extended}, alignment feature-based machine learning methods \cite{sultan2015feature}, etc.The performance of these methods for recognition is not satisfactory because they cannot extract the semantic information of the sentences well. In recent years, deep learning-based methods have been effective in semantic modeling and have achieved good results in many tasks of NLP. Therefore, on the task of sentence matching, the performance of deep learning-based methods has surpassed the earlier methods and it becomes the mainstream sentence matching method. For example, Bowman {\it et al}.\cite{bowman2015large}  first applied LSTM sentence models to the RTE domain by encoding premises and hypotheses through LSTM to obtain sentence vectors. Rocktäschel {\it et al}.\cite{rocktaschel2015reasoning} first introduced the attention mechanism to textual implication recognition by finding the most relevant words in the hypothesis and premise sentences for alignment, which can enhance the blending of the two-sentence vectors. mLSTM model was proposed by WANG {\it et al}.\cite{wang2016learning} on this basis, which focuses on splicing attention weights in the hidden states of the LSTM, focusing on the part of the semantic match between the premise and the hypothesis. The experimental results showed that the method achieved good results on the SNLI dataset.Yang {\it et al}.\cite{yang2019simple} proposed a simple and efficient sentence matching model, which used convolutional neural networks to encode sentence vectors and achieved good results on several tasks.

\subsection{Interaction-based sentence matching model}

The interaction-based sentence matching model focuses on obtaining complex interaction information between sentence pairs.Wang {\it et al}.\cite{wang2017bilateral} proposes the BIMPM model, which first encodes sentence pairs through a bidirectional LSTM, and then matches encoding results from multiple perspectives in two directions, achieving better results in multiple tasks of natural language inference, paraphrase recognition, and answer selection.Chen {\it et al}.\cite{chen2017enhanced} proposes an ESIM model that uses a two-layer bidirectional LSTM and a self-attentive mechanism for coding, and then extracts features through the average pooling layer and the maximum pooling layer, and finally performs classification. Compared to ESIM with larger dimensional vectors, Tay {\it et al}.\cite{tay2018compare} proposes the CAFÉ model, which compresses the feature vectors and achieves a lightweight model. Kim {\it et al}.\cite{kim2019semantic} connected multiple cycles of encoding and interaction modules through residuals for extracting  semantic information, and also solved the gradient vanishing/explosion problem.

The models mentioned above have achieved good results on specific tasks, but most of these models can only perform well on one or two tasks and have weak generalization ability. To solve the above problems,in this paper, we propose a sentence matching model based on deep coding and interactive sentence matching model and experiment on several tasks with better results.
\begin{figure}[h]
\centering
\includegraphics[width=4.6in]{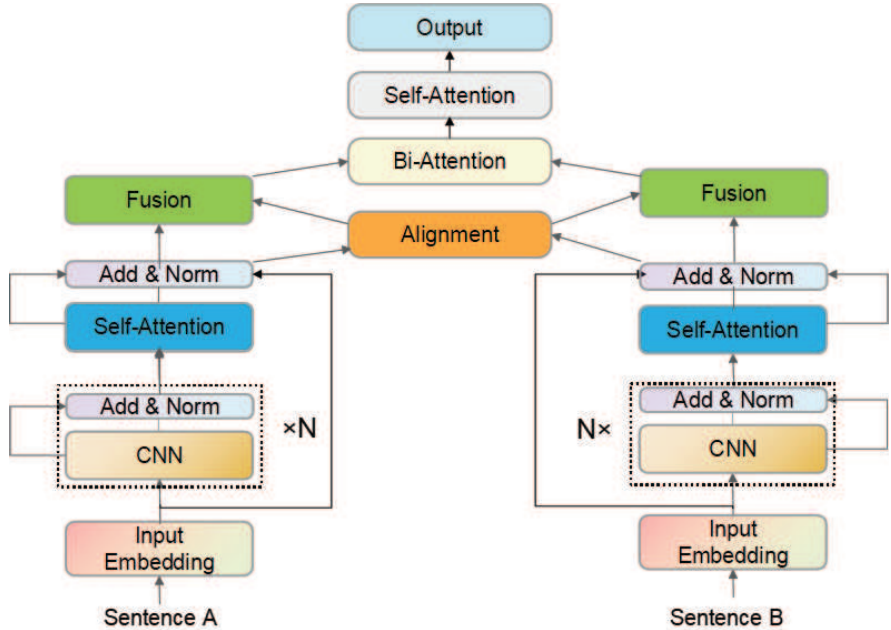}
\caption{Overview of the architecture of our proposed DEIM model.The left part shows the encoding process of the first sentence, and the right part shows the encoding process of the second sentence. The upper module is the interaction process and the output part}
\label{fig:1}
\end{figure}
\section{Method}
In this section,we describe our model in detail. As shown in Figure~\ref{fig:1}, our model  mainly consists of an embedding layer, a contextual encoder layer, a bidirectional attention layer, a self-attention layer, and an output layer.
\subsection{Input Embedding Layer}
The purpose of the embedding layer is to map the input sentence $A$ and sentence $B$ into word vectors.The usual practice is to use a pre-trained word vector Glove for word embedding, however, this word embedding method cannot solve the problem of multiple meanings of words. To solve the above problems, we use a pre-trained static word vector Glove combined with a dynamic word vector Elmo, which is stitched together as a vector representation of words\cite{peters2018deep}.

We assume that the corresponding static word vector for word $h$ is ${h_1}$ , and the dynamic word vector is ${h_2}$ . We splice the two vectors to get the word vector representation of word $h$:$h = [{h_{\rm{1}}};{h_{\rm{2}}}] \in {R^{{d_1} + {d_2}}}$ , where ${d_1}$  is the dimension of Glove word embedding and  ${d_2}$ is the dimension of Elmo word embedding. Finally, we obtain the word embedding matrix $X \in {R^{n{\rm{*}}({d_1} + {d_2})}}$ for sentence $A$ and the word embedding matrix $Y \in {R^{m*({d_1} + {d_2})}}$ for sentence $B$, where $n$,$m$ represent the number of words in sentence $A$ and sentence $B$.

\subsection{Contextual Encoder Layer}
The purpose of the contextual encoder layer is to fully exploit the contextual relationship features of the sentences.Generally, most models use bidirectional LSTM for encoding\cite{wang2017bilateral}\cite{chen2017enhanced}, which can mine the contextual relationship features of the sentences, but there is still the problem of gradient disappearance for long sentence sequences. Meanwhile, since the LSTM adopts the architecture of RNN, 
it has a slow encoding efficiency. Therefore, we use CNN and self-attention for encoding. Local features are first extracted using multiple CNNs in the encoding layer, and then global features are obtained using self-attention. In order to fully integrate the semantic information obtained by encoding,in this paper,we add alignment and fusion modules. This is done as  follows: Firstly,we obtained the word embedding matrix of sentence $A$ and sentence $B$ through embedding layer.Secondly,we  use multiple CNNs as well as Self-Attention to obtain its representation $C \in {R^{n*d}}$ and $Q \in {R^{m*d}}$ , where $d$ is the hidden layer dimension. Next, We use the alignment module to get the aligned sentences representing ${C^{'}}$ and ${Q^{'}}$, the process is shown in Eq.\ref{eq1}:
\begin{equation}\label{eq1}
\begin{aligned}
    & S=relu{{({{W}_{c}}{{C}^{T}})}^{T}}relu({{W}_{q}}{{Q}^{T}}) \\ 
 & a=softmax(S) \\ 
 & {{C}^{'}}=a\bullet Q \\ 
 & {{Q}^{'}}={{a}^{T}}\bullet C \\ 
\end{aligned}
\end{equation}
where ${W_c}$ and ${W_q}$  are the learnable parameters. Finally, we fuse $C$ and ${{C}^{'}}$ to obtain the encoded representation $H=fusion(C,{{C}^{'}})\in {{R}^{n*d}}$ of sentence $A$ , where the fusion function is defined as shown in Eq.\ref{eq2}:
\begin{equation}\label{eq2}
\begin{aligned}
    & \widetilde{x}=\tanh ({{W}_{1}}[x;y;x\odot y;x-y]) \\ 
 & g=sigmoid({{W}_{2}}[x;y;x\odot y;x-y]) \\ 
 & z=g\odot \widetilde{x}+(1-g)\odot x \\ 
\end{aligned}
\end{equation}
Where ${{W}_{1}}$ and ${{W}_{2}}$  are weight matrices, and $g$ is a gating mechanism to control the weight of the intermediate vectors in the output vector. In this paper, $x$ refers to $C$ and $y$ refers to ${{C}^{'}}$.Similarly, the final encoding representation $P\in {{R}^{m*d}}$ of sentence $B$ can be obtained.
\subsection{Bidirectional Attention Layer}
The purpose of the bidirectional attention layer is to calculate the bidirectional attention of $H$ and $P$ that is the attention of $H\to P$ and $P\to H$\cite{seo2016bidirectional} . The attention  originates from the similarity matrix \bm{$S$} obtained in the previous layer, where ${{S}_{ij}}$ denotes the similarity between the $i$-th word of $H$ and the $j$-th word of $P$.

$H\to P$: The attention describes which words in the sentence $P$ are most relevant to $H$. The calculation process is as follows; first, each row of the similarity matrix is normalized to get the attention weight, and then the new text representation $Q\in {{R}^{2d*n}}$ is obtained by weighted summation with each column of $P$,which is calculated as shown in Eq.\ref{eq3}.
\begin{equation}\label{eq3}
\begin{array}{l}
	\alpha _t=soft\max \left( S_{t:} \right) \in R^m\\
	q_{:t}=\sum_j{\alpha _{tj}P_{:j}}\\
\end{array}
\end{equation}

where ${q_{:t}}$ is the $t$-th column of Q.

$P\to H$: The attention indicates which words in $H$ are most similar to $P$. The calculation process is as follows: first, the column with the largest value in the similarity matrix \bm{$S$} is taken to obtain the attention weight, then the weighted sum of $H$ is expanded by $n$ time steps to obtain $C\in {{R}^{2d*n}}$,which is calculated as shown in Eq.\ref{eq4}.
\begin{equation}\label{eq4}
\begin{aligned}
    & b=softmax (\underset{col}{\mathop{\max }}\,(S))\in {{R}^{n}} \\ 
 & c=\sum\limits_{t}{{{b}_{t}}{{H}_{t:}}\in {{R}^{2d}}} \\ 
\end{aligned}
\end{equation}

After obtaining the attention matrix $Q$ of $H\to P$and the attention matrix $C$ of $P\to H$, we splice the attention in these two directions by a multilayer perceptron. Finally, we get the spliced contextual representation $G$, which is calculated as shown in Eq.\ref{eq5}.
\begin{equation}\label{eq5}
\begin{aligned}
    & {{G}_{:t}}=\beta ({{C}_{:t}},{{H}_{:t}},{{Q}_{:t}}) \\ 
 & \beta (c,h,q)=[h;q;h\odot q;h\odot c]\in {{R}^{8d}}  
\end{aligned}
\end{equation}
\subsection{Self-attention Layer}
The purpose of the self-attention layer is to extract the deep semantics of the text and provide a deep understanding of the text information. The matrix $G$ that fuses bidirectional attention is obtained in the previous layer although it can fully fuse the semantic information between $H$ and $P$, it does not consider the information about itself. Therefore, in this layer, we capture the deep semantic information of the text through the self-attention mechanism \cite{vaswani2017attention}.  We calculate the text-to-text similarity matrix $E\in {{R}^{n*n}}$ . It is row normalized to obtain the attention weights. We can obtain the self-attention representation $Z$ of the text. The calculation formula is shown in Eq.\ref{eq6}.
\begin{equation}\label{eq6}
\begin{aligned}
    & E={{G}^{T}}G \\ 
 & Z=G\cdot softmax (E) 
\end{aligned}
\end{equation}

\subsection{Output Layer}

The purpose of the output layer is to output the results. In this paper, we use a linear layer to get the results of text matching. The process is shown  in Eq.\ref{eq7}.
\begin{equation}\label{eq7}
\begin{aligned}
   y = F(\tanh (ZW + b))
\end{aligned}
\end{equation}
where both $W$ and $b$ are trainable parameters. $Z$ is the vector after splicing its first and last vectors. Generally,for the sorting task, the activation function of $F$ is $tanh$; for the classification task, the activation function of $F$ is $softmax$.
\section{Experiments}
In this section,we validate our model on four datasets from three tasks. We first present some details of the model implementation, and secondly, we show the experimental results on the dataset. Finally, we analyze the experimental results.
\subsection{Experimental details}
\subsubsection{Loss function}
For the classification task, the cross-entropy loss function can be chosen as shown  in Eq.\ref{eq8}.
\begin{equation}\label{eq8}
\begin{aligned}
   loss=-\sum\limits_{i=1}^{N}{\sum\limits_{k=1}^{K}{{{y}^{(i,k)}}\log {{{\hat{y}}}^{(i,k)}}}}
\end{aligned}
\end{equation}
where $N$ is the number of samples, $K$ is the total number of categories and ${{\hat{y}}^{(i,k)}}$ is the true label of the $i$-th sample. For the sorting task, this paper uses Hinge Loss, as shown in Eq.\ref{eq9}.
\begin{equation}\label{eq9}
\begin{aligned}
   loss(A,{{B}^{+}},{{B}^{-}})=\max (0,1-f(A,{{B}^{+}})+f(A,{{B}^{-}}))
\end{aligned}
\end{equation}
where $B{}^{+}$ and $B{}^{-}$ are the samples related/unrelated to $A$ respectively; $f(A,B)$ denotes the matching score obtained by the model.
\subsubsection{Dataset}
\begin{figure}
\centering
\includegraphics[width=2.6in]{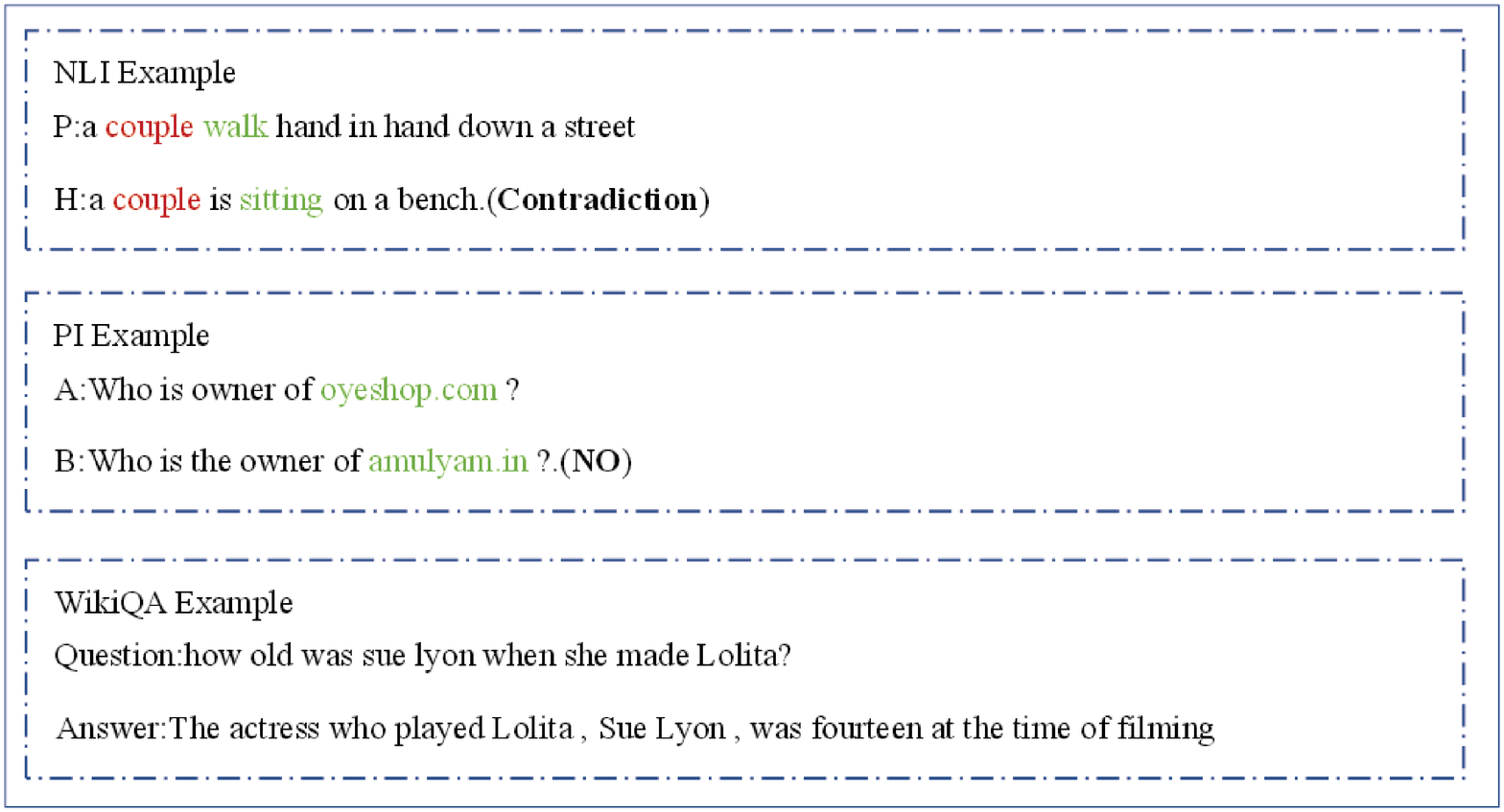}
\caption{dataset case}
\label{fig:2}
\end{figure}

In this paper, we use the natural language inference datasets SNLI and SciTail, the paraphrase recognition dataset Quora, and the answer selection dataset WikiQA to validate our model. Figure~\ref{fig:2} shows a sample of the datasets. Among them, the SNLI dataset contains 570K manually labeled and categorically balanced sentence pairs. The SciTail dataset contains 27k sentence pairs and it is a textual entailment dataset created from multiple-choice scientific quiz tasks and web sentences,it  contains implied and neutral labels.The Quora question pair dataset contains over 400k pairs of  data that each with binary annotations, with 1 being a duplicate and 0 being a non-duplicate. The WikiQA dataset is an answer selection dataset based on Wikipedia, which contains more than 1,200 questions with corresponding candidate answers.  The statistical descriptions  of SNLI, SciTail, Quora, and WikiQA data are shown in Table \ref{tab:1}.
\begin{table}[htbp]
 \centering
  \caption{\label{tab:1}The statistical descriptions of SNLI, Scitail, Quora, and WikiQA}
\begin{tabular}{cccc}
 \toprule
dataset & train & validation  & test\\
 \midrule
SNLI & 550152 & 10000 & 10000 \\
SciTail & 23596 & 1304 & 2126 \\
Quora & 384290 & 10000 & 10000 \\
WikiQA & 20360 & 2733 & 6165 \\
\bottomrule
 \end{tabular}
\end{table}
\subsubsection{Baseline methods and parameter settings}
We compare with thirteen baseline methods on the dataset, including  representation-based models (i.e., RE2 (Yang et al.),DITM(Yu et al.),DELTA(Han et al.),BiLSTM(Borges et al.),Transformer(Guo et al.;Radford et al.) ), interaction-based models (i.e., DRr-Net (Zhang et al.), DRCN (Kim et al.), HCRN(Tay et al.),DITM(Yu et al.),ESIM (Chen et al.) , BiMPM (Wang et al.) and SWEM-max(Shen et al.)) .

This experiment is conducted in a hardware environment with a graphics card RTX5000 and 16G of video memory. The system is Ubuntu 20.04, the development language is Python 3.7, and the deep learning framework is Pytorch 1.8.

In the model training process, a 300-dimensional Glove word vector and a 1024-dimensional ELMO word vector are used for word embedding, and the maximum length of text sentences is set to 64, 48, 48, and 32 words on the SNLI, SciTail, Quora, and WikiQA datasets, respectively. The specific hyperparameter settings are shown in Table \ref{tab:2}.
\begin{table}[htbp]
 \centering
  \caption{\label{tab:2}Values of Hyper Parameters}
\begin{tabular}{cc}
 \toprule
Hyper Parameters & Values\\
 \midrule
Glove dimension & 300 \\
Elmo dimension & 1024 \\
hidden dimension & 150 \\
convolution kernel size & 3 \\
learning rate & 0.0005 \\
Optimizer & Adam \\
Dropout & 0.2 \\
activation function & ReLU \\
Epoch & 30 \\
Batch size & 128 \\
\bottomrule
 \end{tabular}
\end{table}
\subsection{Experimental results and analysis}
We compare the experimental results of the deep encoding and interaction-based sentence matching model on the SNLI dataset with other published models.The evaluation metric we use is the accuracy rate. The results are shown in Table 3. As can be seen from Table \ref{tab:3}, our model achieves an accuracy rate of 0.889 on the SNLI dataset, which achieves better results in the listed models. Compared with the bidirectional LSTM, it is improved by 0.044. Compared with some other models, 
it is observed that the latter is better than the former.
\begin{table}[htbp]
 \centering
  \caption{\label{tab:3}The accuracy($\%$) of the model on the SNLI test  set.Results marked with    $^a$ are reported by Han et al.\cite{han2019delta},  $^b$ are reported by Shen et al.\cite{shen2018baseline}, $^c$ are reported by Borges et al.\cite{borges2019combining}, $^d$ are reported by Guo et al.\cite{guo2019star}, $^e$ are reported by Wang et al.\cite{wang2017bilateral}, $^f$ are reported by Chen et al.\cite{chen2017enhanced}, $^g$ are reported by Yang et al.\cite{yang2019simple}.}
\begin{tabular}{cc}
 \toprule
Model & Acc\\
 \midrule
DELTA$^a$ & 80.7 \\
SWEM-max$^b$ & 83.8 \\
Stacked Bi-LSTMs$^c$ & 84.8 \\
Bi-LSTM sentence encoder$^c$ & 84.5 \\
Star-Transformer$^d$ & 86.0 \\
BIMPM$^e$ & 87.9 \\
ESIM$^f$ & 88.0 \\
\textbf{RE2}$^g$ & \textbf{88.9} \\
\textbf{DEIM} & \textbf{88.9} \\
\bottomrule
 \end{tabular}
\end{table}
We conduct experiments on the SciTail dataset, and the evaluation metric is accuracy.
The results of our experiments on the Scitail dataset are shown in Table \ref{tab:4}. Our method achieves an accuracy of 0.895 ,which achieves better results in comparison with other models. Compared with the classical ESIM model, it is improved by 0.189. Also,compared with some other models, 
the accuracy of most of our models has also improved.  
\begin{table}[htbp]
 \centering
  \caption{\label{tab:4}The accuracy($\%$) of the model on the SciTail test  set.Results marked with    $^f$ are reported by Chen et al.\cite{chen2017enhanced}, $^g$ are reported by Yang et al.\cite{yang2019simple},$^h$ are reported by Zhang et al.\cite{zhang2019drr}, $^i$ are reported by Radford et al.\cite{radford2018improving}.}
\begin{tabular}{cc}
 \toprule
Model & Acc\\
 \midrule
ESIM$^f$ & 70.6 \\
RE2$^g$ & 86.6 \\
DRr-Net$^h$ & 87.4 \\
Transformer LM$^i$ & 88.3 \\
\textbf{DEIM} & \textbf{89.5} \\
\bottomrule
 \end{tabular}
\end{table}
We conduct experiments on the Quora dataset, and the evaluation metric is accuracy.
The experimental results on the Quora dataset are shown in Table  \ref{tab:5}. As can be seen from Table \ref{tab:5}, the accuracy of our method on the test set is 0.901. The experimental results improve the accuracy by 0.087 compared to the traditional LSTM model, 0.047 compared to the enhanced sequential inference model ESIM, and 0.007 compared to the RE2 model with multiple interactions. 
\begin{table}[htbp]
 \centering
  \caption{\label{tab:5}The accuracy($\%$) of the model on the Quora test  set.Results marked with    $^j$ are reported by Zhao et al.\cite{zhao2020algorithm}, $^k$ are reported by Kim et al.\cite{kim2019semantic} , $^e$ are reported by Wang et al.\cite{wang2017bilateral},  $^f$ are reported by Chen et al.\cite{chen2017enhanced}, $^g$ are reported by Yang et al.\cite{yang2019simple}.}
\begin{tabular}{cc}
 \toprule
Model & Acc\\
 \midrule
LSTM & 81.4 \\
Capsule-BiGRU$^j$ & 86.1 \\
ESIM$^f$ & 85.4 \\
BIMPM$^e$ & 88.2 \\
RE2$^g$ & 89.4 \\
\textbf{DRCN}$^k$ & \textbf{90.1} \\
\textbf{DEIM} & \textbf{90.1} \\
\bottomrule
 \end{tabular}
\end{table}
We conduct experiments on the WikiQA dataset, and the evaluation metric is MAP and MRR.
The experimental results on the WikiQA dataset are shown in Table \ref{tab:6}. Our model achieves better results (MAP and MRR values of 0.755 and 0.775, respectively), with MAP and MRR improving by 0.103 and 0.111 compared to the enhanced sequential inference model ESIM, and by 0.037 and 0.044 compared to the bidirectional multi-view matching model BIMPM. when we compare with other models, we still get the best results. Our models have good performance on different tasks, which illustrates the effectiveness of our model.
\begin{table}[htbp]
 \centering
  \caption{\label{tab:6}The MAP($\%$) and MRR($\%$) of the model on the WikiQA test set.Results marked with    $^l$ are reported by Tay et al.\cite{tay2018hermitian}, $^m$ are reported by Yu et al.\cite{yu2021simple} , $^e$ are reported by Wang et al.\cite{wang2017bilateral},  $^f$ are reported by Chen et al.\cite{chen2017enhanced}, $^g$ are reported by Yang et al.\cite{yang2019simple}}
\begin{tabular}{ccc}
 \toprule
Model & MAP & MRR\\
 \midrule

ESIM$^f$ & 65.2 & 66.4 \\
BIMPM$^e$ & 71.8 & 73.1 \\
HCRN$^l$ & 74.3 & 76.2 \\
RE2$^g$ & 74.5 & 76.2 \\
DITM$^m$ & 75.2 & 77.0 \\
\textbf{DEIM} & \textbf{75.5}&  \textbf{77.5} \\
\bottomrule
 \end{tabular}
\end{table}
\subsection{ Ablation experiments}
To explore the role played by each module, we conduct an ablation experiment  on the SNLI dataset . Without using the fusion function, which means that the original vectors are directly spliced with the vectors after alignment. The experimental results are shown in Table \ref{tab:7}.
\begin{table}
 \centering
  \caption{\label{tab:7}Ablation study  on the SNLI validation dataset}
\begin{tabular}{cc}
 \toprule
Model & Acc($\%$)\\
 \midrule
DEIM & 88.9 \\
w/o ELMO & 87.8 ($\downarrow $1.1)\\
w/o alignment & 87.5(\textbf{$\downarrow $1.4}) \\
w/o fusion & 88.0($\downarrow $0.9) \\
w/o self-attention & 88.2($\downarrow $0.7) \\
Only $H\to P$ & 87.9($\downarrow $1.0) \\
Only $P\to H$ & 88.1($\downarrow $0.8) \\
\bottomrule
 \end{tabular}
\end{table}
We first verify the effectiveness of dynamic word embedding. Specifically,we remove the dynamic word vector for the experiment, and its accuracy drops by 1.1 percentage points, proving that dynamic word embedding plays an important role in improving the performance of the model.

In addition,we verify the effectiveness of the sentence alignment and fusion modules. We removed the sentence soft alignment module from the original model, and its accuracy dropped by 1.4 percentage points. At the same time, we remove the fusion function, and its accuracy drops by about 0.9 percentage points. It shows that the sentence soft alignment module and the fusion function are beneficial to improve the accuracy of the model.

Finally,we verify the effectiveness of each attention on the model.We remove the attention from $H$ to $P$, the attention from $P$ to $H$, and the self-attention module respectively. Their accuracy rates decreased by 0.8 percentage points, 1.0 percentage points, and 0.7 percentage points. It shows that all the various attention mechanisms improve the performance of the model, with the $P$ to $H$ attention being more significant for the model.

The ablation experiments show that each component of our model plays an important role, especially the alignment module and the word embedding module, which have a greater impact on the performance of the model.
\section{Conclusion}
we investigate natural language sentence matching methods and propose a sentence matching method based on deep coding and interaction. Our model is encoded using a deep encoder, which includes an alignment module and a fusion module. After that,we use bidirectional attention flow for interaction  and  the self-attention mechanism  to obtain deep information . We conducted experiments on SNLI, SciTail, Quora, and WikiQA datasets. The experimental results show that the model proposed in this paper can achieve better results in all three tasks.In this work, we find that our proposed encoder layer occupies the most dominant position and has a greater impact on our model. It also indirectly shows the effectiveness of the encoding.However, 
our model lacks external knowledge. External knowledge plays an important role for sentence matching and can achieve good results in the case of sparse data. The introduction of external knowledge will be considered in future work. For example, WordNet, an external knowledge base, contains many sets of synonyms, and for each input word, its synonyms are retrieved from WordNet and embedded in the word vector representation of the word to further improve the performance of the model.
\section*{Acknowledgement}
This work is supported by National Natural Science Foundation of China [grant numbers 62162062], State Language Commission of China under Grant No. YB135-76, scientific research project for building world top discipline of Foreign Languages and Literatures of Yanbian University under Grant No. 18YLPY13, Doctor Starting Grants of Yanbian University [2020-16], the school-enterprise cooperation project of Yanbian University [2020-15].
\section*{}

\end{document}